\documentclass[11pt, oneside]{article}   	% use "amsart" instead of "article" for AMSLaTeX format
\usepackage{geometry}                		% See geometry.pdf to learn the layout options. There are lots.
\usepackage{graphicx}				% Use pdf, png, jpg, or eps§ with pdflatex; use eps in DVI mode
\usepackage{amssymb}

\usepackage{times}  %Required
\usepackage{helvet}  %Required
\usepackage{courier}  %Required
\usepackage{url}  %Required
\usepackage{graphicx}  %Required
\usepackage{amsfonts}
\usepackage{mathtools,amssymb}
\usepackage{amsmath}
\usepackage{algorithm}
\usepackage{algorithmic}
\usepackage{subfigure}

\DeclareMathOperator*{\argmin}{arg\,min}

\newtheorem{definition}{Definition}
\newtheorem{problem}{Problem}
\newtheorem{theorem}{Theorem}

\title{Latent Space Optimal Transport for Generative Models}
\author{Huidong Liu,$^1$ Yang Guo,$^1$ Na Lei,$^2$ Zhixin Shu,$^1$ \\ Shing-Tung Yau,$^3$ Dimitris Samaras,$^1$ Xianfeng Gu$^1$ \\
$^1$Department of Computer Science, Stony Brook University, USA \\
$^2$School of Software Technology, Dalian University of Technology, China\\
$^3$Harvard University, USA \\
\texttt{\{huidliu, yangguo, samaras, gu\}@cs.stonybrook.edu} \\
\texttt{nalei@dlut.edu.cn} \\
\texttt{yau@math.harvard.edu}}

\date{}							% Activate to display a given date or no date

\begin{document}
\maketitle
\begin{abstract}
Variational Auto-Encoders enforce their learned intermediate latent-space data distribution to be a simple distribution, such as an isotropic Gaussian. However, this causes the posterior collapse problem and loses manifold structure which can be important for datasets such as facial images. A GAN can transform a simple distribution to a latent-space data distribution and thus preserve the manifold structure, but optimizing a GAN involves solving a Min-Max optimization problem, which is difficult and not well understood so far. Therefore, we propose a GAN-like method to transform a simple distribution to a data distribution in the latent space by solving only a minimization problem. This minimization problem comes from training a discriminator between a simple distribution and a latent-space data distribution. Then, we can explicitly formulate an Optimal Transport (OT) problem that computes the desired mapping between the two distributions. This means that we can transform a distribution without solving the difficult Min-Max optimization problem. Experimental results on an eight-Gaussian dataset show that the proposed OT can handle multi-cluster distributions. Results on the MNIST and the CelebA datasets validate the effectiveness of the proposed method. 
\end{abstract}

\section{Introduction}
Auto-Encoders (AEs) have demonstrated the capability of learning a subspace for dimensionality reduction \cite{liou2008modeling}. However, AEs are not generative. Mathematically speaking, there exist zones where a latent code is not in the support of the latent representation of the input \cite{tolstikhin2017wasserstein}. In order to address this problem, Variational Auto-Encoders (VAEs) \cite{kingma2013auto} enforce the latent-space data distribution to be close to a simple distribution, e.g., a unit Gaussian, such that a randomly sampled latent code lies in the support of the latent representation of the given dataset. In practice, VAEs minimize the KL-divergence between the latent-space data distribution and a unit Gaussian \cite{kullback1951information}. For a similar purpose, instead of measuring the KL-divergence, the Adversarial Auto-Encoder (AAE) \cite{makhzani2015adversarial} adopts adversarial training in the latent space to enforce the latent-space data distribution to be a unit Gaussian \cite{berthelot2017began}. The Wasserstein Auto-Encoder (WAE) \cite{tolstikhin2017wasserstein} with a GAN penalty (WAE-GAN) is the generalization of the AAE with the reconstruction cost being any cost function. When the cost function is quadratic, WAE-GAN reproduces AAE. 

Existing VAE based methods \cite{mescheder2017adversarial,larsen2015autoencoding}, AAE and WAE transform the distribution of the latent representations of data to a simple distribution such as an isotropic Gaussian. However, many real world datasets, such as facial images, lie in lower dimensional manifolds which are quite different from a simple isotropic Gaussian \cite{he2005face,arandjelovic2009unfolding}. Changing the important manifold structure of data into a simple distribution, e.g. a unit Gaussian, obliterates the structure of latent-space data distribution, causes the posterior collapse problem \cite{van2017neural} and leads to generating unrealistic images. %Applying VAE on these data ruins the intrinsic manifold structure of data in the latent space.
% \emph{The transformation between distributions in the latent space is obscure: hidden in the training process implicitly. The transformation quality has no theoretical guarantee.}

In contrast to VAE-based methods in which the distribution of the latent representation of data is transformed to an isotropic Gaussian implicitly using KL-divergence, Generative Adversarial Nets (GANs) \cite{goodfellow2014generative} are able to transform any given distribution to another distribution theoretically. GAN based methods are receiving increased attention in various applications \cite{NeuralFace2017,VuICCV2017,zhu2017unpaired,li2017generative,iizuka2017globally,hou2017unsupervised} as well as methodology improvements \cite{salimans2016improved,arjovsky2017wasserstein,gulrajani2017improved,miyato2018spectral,mirza2014conditional}. A GAN model consists of a generator and a discriminator (or critic): the generator synthesizes data from a simple distribution to fool the discriminator while the discriminator tries to distinguish between the real data and synthetic data. However, training a GAN is solving a Min-Max optimization problem \cite{radford2015unsupervised}, which is difficult and unstable in practice \cite{mescheder2017adversarial}. Furthermore, the balance between the discriminator and generator is difficult to control \cite{berthelot2017began}. 

\begin{figure} [t]
\centering     %%% not \center

\subfigure[The workflow of AE-OT]{\label{fig:subfig:AE_OT_workflow}\includegraphics[width=70mm]{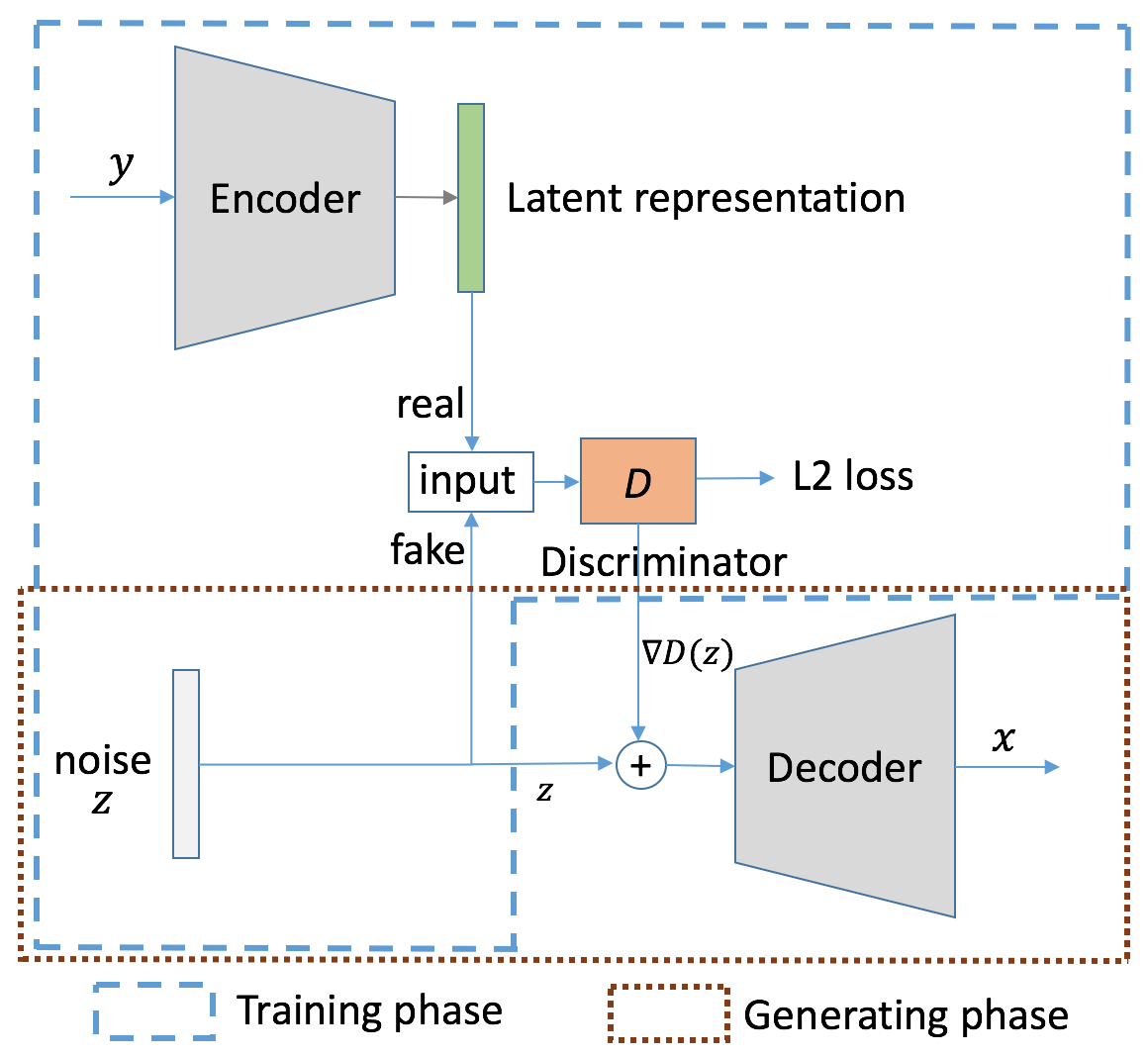}}
\caption{The workflow of the AE-OT. The Encoder and Decoder are trained and then fixed. In the training phase, the latent representation of a sample $y$ is real data for the discriminator, while a noise sample $z$ is fake data. We train the discriminator $D$ using the proposed approach. In the generating phase, we generate a noise sample $z$ and input $z + \nabla D(z)$ into the Decoder to produce a generated data sample $x$. }
\label{fig: workflow}
\end{figure}

In this paper, we address the posterior collapse problem in VAE by learning a transformation from a simple distribution (e.g. a unit Gaussian) to the latent-space data distribution. This preserves the structure of the data in the latent space. Traditionally, in order to achieve this goal, a GAN with two networks (a generator and a discriminator) is used. However, the adversarial
training process is not well understood theoretically. In contrast, our proposed method computes the Optimal Transport (OT) map directly based on the theoretical analysis presented in \cite{lei2017geometric}. We only train a discriminator in the latent space and the OT which transforms a sample from a simple distribution to a latent-space data distribution is explicitly derived from the discriminator output. As we use an Auto-Encoder (AE) to find the latent space and use OT to perform distribution transformation in the latent space, we name our method as AE-OT. The OT has a well understood theory, and hence we can use transparent theoretic model to transform a distribution. Figure \ref{fig: workflow} shows the workflow of AE-OT.

In contrast to the Wasserstein Auto-Encoder (WAE) in which the Wasserstein Distance (WD) is defined in the original image space, in AE-OT, the WD is defined in the latent space. Both the objective and the training protocol are different.
WAE requires solving a difficult Min-Max optimization problem, while AE-OT only needs to solve a minimization problem in the latent space. So, training AE-OT is much easier than training WAE. It is worthy noting that several OT methods have been proposed in literature. \cite{stavropoulou2015parametrization,ferradans2014regularized} compute discrete OT from OT primal formulation, and thus are not suitable for generative models. \cite{perrot2016mapping} learns the OT mapping using kernel methods, whose parameters are difficult to choose. \cite{seguy2017large} needs to train three networks to learn the transport mapping, which is harder than AE-OT. 

The contributions of this paper are the following:

1) We propose a novel generative Auto-Encoder. Different from existing VAE based methods which map a latent-space data distribution to a simple distribution, the proposed generative model transforms a simple distribution to the latent-space data distribution. In this way, the intrinsic data structure in the latent space is preserved, and thus the posterior collapse problem \cite{van2017neural} is addressed. 
%This method is based on the solid optimal transportation theory. Thus it can be easily interpreted and has theoretical advantages.
%with explicit theoretic interpretation and optimal convergence guarantees. 

2) We show that if the cost function is quadratic, then once the optimal discriminator is obtained, the generator can be explicitly obtained from the discriminator output. AE-OT can achieve the same goal of a GAN in the latent space, but AE-OT only needs to solve a minimization problem in the latent space rather than the difficult Min-Max optimization problem in GANs. 

3) Experiments on an eight-Gaussian toy dataset demonstrate that the computed OT can model the multi-cluster distributions. Qualitative and quantitative results on the MNIST \cite{lecun1998gradient} dataset show that AE-OT performs  better than VAEs and WAE. Images generated on the CelebA \cite{liu2015faceattributes} dataset show that AE-OT generates much better facial images than VAE and WAE. 

In this remainder of this paper, we shall review optimal transport first, and then introduce our proposed generative model, followed by experimental results. 

\section{Optimal Transport}
\label{sec_theory}

Since our method is based on Optimal Transport (OT), we  first introduce the background of OT. OT is a powerful tool to handle probability measure transformations. For details one could refer to \cite{lei2017geometric} \cite{villani2008optimal}.
\subsection{Optimal Transport theory}
In this subsection, we will introduce basic concepts and theorems in classic optimal transport theory, focusing on Kantorovich potential and Brenier's approach of solving the Monge problem defined in \textbf{Problem} \ref{prob: Monge}. 

Let $X\subset \mathbb{R}^d$, $Y\subset \mathbb{R}^d$ be two subsets of a $d$-dimensional Euclidean space, with measures $\mu$ and $\nu$ be probability measures defined on $X$ and $Y$, respectively. Also we require that they have equal total measure, i.e
\[
    \int_X d\mu = \int_Y d\nu.
\]
\begin{definition}[Measure-Preserving Map]
A map $T: X\to Y$ is \emph{measure preserving} if for any measurable set $B\subset Y$, the set $T^{-1}(B)$ is $\mu$-measurable and 
\begin{equation}
    \mu(T^{-1}(B)) = \nu(B).
    \label{eqn:area_preserving_mapping}
\end{equation}

\end{definition}
The problem of optimal transport arises from minimizing the total cost of moving all particles from one place (i.e. source) to another place (i.e. target), given the cost of moving each unit of mass. Formally speaking, we define a cost function $c(x,y)$ on $X\times Y$, such that $c(x,y) \ge 0$ for every $x\in X$ and $y\in Y$. The total {\it transport cost} of moving particles with density $\mu$ at $X$ to density $\nu$ at $Y$ is defined to be 
\begin{equation}
    \int_X c(x,T(x)) d\mu(x).
    \label{eqn:transportation_cost}
\end{equation}
Eq. (\ref{eqn:transportation_cost}) can also be rewritten as $T_\#\mu = \nu$, where $T_\#$ is the push forward map induced by $T$. 
Now we could define the Monge's problem of optimal transport. 

\begin{problem}
\label{prob: Monge}
[Monge's Optimal Transport \cite{bonnotte2013knothe}] Given a transport cost function $c: X\times Y\to \mathbb{R}$, find a measure preserving map $T:X\to Y$ that minimizes the total transport cost
\begin{equation}
(MP) \quad \min_{T:X\to Y} \left\{\int_X c(x,T(x)) d\mu(x):T_\#\mu=\nu \right\}.
    \label{eqn:MP}
\end{equation}
\end{problem}

Note that in Monge's problem a map $T: X\mapsto Y$ is required, whose existence is not guaranteed given an arbitrary pair of measures $\mu$ and $\nu$. For example when $\mu$ is a Dirac measure and $\nu$ is an arbitrary measure absolutely continuous to Lebesgue measure in $\mathbb{R}^d$. This means that given $\mu$ and $\nu$, the feasible solution set to Monge's problem might be empty. Therefore, Kontorovich introduced a relaxation of Monge's problem \cite{kantorovich2006problem} in 1940s. Instead of considering a transport map, he considered the set of {\it transport plan}. Mathematically, given source measure $\mu$ on $X$ and target measure $\nu$ on $Y$, a transport plan is a joint distribution $\rho$ on $X\times Y$, such that
\begin{equation}
    \rho(A\times Y) = \mu(A),\ \rho(X\times B) = \nu(B),
    \label{eqn:marginal_measure}
\end{equation}
Intuitively, if $\rho(A\times B) > 0$ for some set $A\subset X$, $B\subset Y$, there will be mass moved from $A$ to $B$ in plan $\rho$.  Now the total cost of transport plan $\rho$ is 
\begin{equation}
    \mathcal{C}(\rho):= \int_{X\times Y} c(x,y) d\rho(x,y).
    \label{eqn:transportation_cost_2}
\end{equation}

And the Monge-Kantorovich problem is defined as 
\begin{equation}
\begin{split}
(KP) \quad \min_{\rho} & \quad \left\{ \int_{X\times Y} c(x,y) d\rho(x,y) \right\} \\
\mbox{s.t.} &\quad \pi_{x\#}\rho = \mu,~\pi_{y\#}\rho = \nu
    \label{eqn:KP}
\end{split}
\end{equation}
among all transport plans $\rho$, where $\pi_{x}$ and $\pi_{y}$ are projection maps from $X\times Y$ onto $X$ and $Y$ respectively.

To solve the ($KP$) problem, we consider its dual form, known as the Kantorovich problem \cite{villani2008optimal},
\begin{equation}
\label{eqn:dual_formulation}
\begin{split}
(DP)\quad \max_{\varphi,\psi} &\quad \left\{\int_Y \psi(y) d\nu(y) - \int_{X} \varphi(x) d\mu(x) \right\} \\
\mbox{s.t.} &\quad \psi(y) - \varphi(x) \le c(x,y)
\end{split}
\end{equation}

\noindent where $\varphi:X\to\mathbb{R}$ and $\psi:Y\to\mathbb{R}$ are real functions defined on $X$ and $Y$. One of the key observations to solve (DP) is based on the concept of {\it c-transform}. 
\begin{definition}[c-transform] Given a real function $\varphi: X\to \mathbb{R}$, the c-transform of $\varphi$ is defined by
\[
    \varphi^c(y) = \inf_{x\in X} \left(c(x,y) + \varphi(x)\right).
\]
\end{definition}
It can be shown \cite{villani2008optimal} that by replacing $\psi$ with $\varphi^c$ in (\ref{eqn:dual_formulation}), the value of the energy to be maximized will not decrease. Therefore we could just search for the optimal $\varphi$ to solve the Kantorovich problem:
\begin{equation}
(DP) \quad \max_{\varphi} \left\{\int_Y \varphi^c(y) d\nu(y) - \int_{X} \varphi(x) d\mu(x)\right\},
    \label{eqn:DP}
\end{equation}
Here $\varphi$ is called {\it Kantorovich potential}.

Formula (\ref{eqn:DP}) can be rewritten into simpler forms, if we narrow down the choice of  cost functions. For example, if we choose the cost function $c(x,y) = ||x-y||_1$, i.e the $L^1$ distance, then the $c$-transform has the property $\varphi^c = \varphi$, given $\varphi$ being $1$-Lipschitz \cite{villani2008optimal}. And (\ref{eqn:DP}) becomes
\begin{equation}
    \max_{\varphi} \left\{ \int_Y \varphi(y) d\nu(y) - \int_{X} \varphi(x) d\mu(x) \right\},
    \label{eqn:L1DP}
\end{equation}

However, the Kantorovich potential $\varphi$ is usually parameterized by a Deep Neural Network (DNN). To restrict a DNN to be 1-Lipschitz is very difficult \cite{arjovsky2017wasserstein}. Our method adopts the $L^2$ distance as the cost function, because in this case once we computed the optimal Kontorovich potential, the transport map can be written down in explicit form \cite{lei2017geometric}. Since the Brenier potential is the transport map (corresponding to the generator in GANs), we shall introduce Brenier theorem \cite{brenier1991polar} below. Suppose $u:X\to \mathbb{R}$ is a second order continuous convex function. Its gradient map $\nabla u: X\to Y $ is defined as $x\mapsto \nabla u(x).$
\begin{theorem}[Brenier\cite{brenier1991polar}] Suppose $X$ and $Y$ are the Euclidean space $\mathbb{R}^d$, and the transport cost is the quadratic Euclidean distance $c(x,y) = ||x-y||^2$. If $\mu$ is absolutely continuous and $\mu$ and $\nu$ have finite second order moments, then there exists a convex function $u: X\to \mathbb{R}$, whose gradient map $ \nabla u$ gives the solution to the Monge's problem, where $u$ is called the Brenier potential. Furthermore, the optimal transport map is unique.
\label{thm:Brenier}
\end{theorem}
Here $u:X\to\mathbb{R}$ is called the \emph{Brenier potential}. In GANs, the discriminator is served as the Kantorovich potential, and the generator is served as the Brenier potential. The following theorem \cite{lei2017geometric} establishes the relationship between the Brenier potential and the Kantorovich potential:
\begin{theorem}Given $\mu$ and $\nu$ on a compact domain $\Omega\subset \mathbb{R}^d$ there exists an optimal transport plan $\gamma$ for the cost $c(x,y)=h(x-y)$ with $h$ being strictly convex. It is unique and of the form $(id,T_\#)\mu$, provided $\mu$ is absolutely continuous and $\partial \Omega$ is negligible. Moreover, there exists a Kantorovich potential $\varphi$, and $T$ can be represented as
\[
    T(x)= x + (\nabla h)^{-1}\nabla \varphi(x).
\]
\label{thm:DG_relation}
In particular, if we choose $h(z) = \frac{1}{2}||z||^2$, then 
\begin{equation}
\label{eqn:T}
    T(x)=x + \nabla \varphi(x) = \nabla \left( \frac{||x||^2}{2} + \varphi(x) \right) = \nabla u(x).
\end{equation}
\end{theorem}

The above theorem shows that  when the discriminator is optimal, the generator can be directly computed from the discriminator. We will employ this important property in our generative model.

\section{The Proposed Generative Model}
Existing VAE based methods try to transform the   latent data distribution to a simple distribution. However, this changes the intrinsic data structure in the latent space and finally leads to the posterior collapse problem \cite{van2017neural}. In our method, we want to preserve the intrinsic data structure in the latent space, but conversely transform a simple distribution to the latent data distribution.

Different
from existing methods, we first pre-train an AE, and
then we fix the AE and compute the OT map to transform a simple distribution
to the latent-space data distribution, such that the intrinsic
data structure is preserved. we name our method as  AE-OT. AE-OT solves the optimal
transport problem in $L^2$ distance.

AE-OT has a strong geometric motivation. A manifold can be arbitrarily complex, and therefore, learning to transform a simple distribution to the complex manifold can be very difficult. There exists a one-to-one mapping from every neighborhood of a manifold to an Euclidean space which is typically of low dimension. One can get an intuition from \cite{lei2017geometric} Figure 1. Thus, we propose to apply an AE to find the low dimensional space, and then perform distribution transformation in the latent space. This is much easier than transforming distributions in the original input space. 

\subsection{Learning the Optimal Transport in Latent Space}
In this part, we propose to transform  the latent space distribution by only training a discriminator. According to Brenier's theorem \ref{thm:Brenier}, the transport map or generator can be explicitly expressed by using the gradient of the optimal discriminator. Training a discriminator is a minimization problem which is much easier than solving a Min-Max optimization problem. As in real applications, we are given the empirical distribution. We give the discrete case of optimal transport below:
\subsubsection{Discrete Case of Optimal Transport}
A generative model can be defined once we find the optimal transport map from a simple distribution to the distribution of real data. To carry out the computational tasks, we introduce basic ideas when the probability measures $\mu$ and $\nu$ are defined on discrete sets. 

Let $\mathcal{I}$ and $\mathcal{J}$ denote the two disjoint sets of indices. Suppose $\hat{X} = \{x_i\}_{i \in \mathcal{I}}$ and $\hat{Y} = \{Y_j\}_{j \in \mathcal{J}}$ are discrete subsets of $\mathbb{R}^d$, and the cost function is defined by $c_{ij} = c(x_i,y_j)$, where $c_{ij} \ge 0$ are positive real numbers. Suppose the source measure $\mu(x_i) = \mu_i$ and the target measure $\nu(y_j) = \nu_j$. A transport plan $\rho$ is a real function that takes values on $\{(x_i,y_j)\ |\ \forall x_i\in \hat{X}, y_j\in \hat{Y}\}$ such that $\rho_{ij} = \rho(x_i,y_j) \ge 0$, $\sum_{i\in\mathcal{I}} \rho_{ij} = \nu_j$ and $\sum_{j\in\mathcal{J}} \rho_{ij} = \mu_i$. We rewrite the total transport cost (\ref{eqn:transportation_cost_2}) as
\begin{equation}
\hat{\mathcal{C}}(\mu,\nu) = \sum_{i\in\mathcal{I}}\sum_{j\in\mathcal{J}} c_{ij}\rho_{ij}.
\end{equation}
The Monge-Kantorovich problem then can be rewritten as 
\begin{equation}
\begin{aligned}
\label{eqn:dis_kp}
& \min_{\rho} \quad \sum_{i\in\mathcal{I}}\sum_{j\in\mathcal{J}} c_{ij}\rho_{ij} \\
& \mbox{s.t.} \quad \sum_{i\in\mathcal{I}} \rho_{ij} = \nu_j, ~ \forall j\in \mathcal{J} \\
& \quad \quad \sum_{j\in\mathcal{J}} \rho_{ij} = \mu_i, ~ \forall i \in \mathcal{I} \\
&  \quad \quad  \rho_{ij} \ge 0, ~ \forall i \in \mathcal{I},~\forall j\in \mathcal{J}
\end{aligned}
\end{equation}
The Monge-Kantorovich dual problem, which in this case is actually the dual form of (\ref{eqn:dis_kp}), is 
\begin{equation}
\begin{aligned}
\label{eqn:dis_dp}
&\max_{\varphi,\psi} \quad \sum_{j\in \mathcal{J}} \psi_j \nu_j - \sum_{i\in \mathcal{I}} \varphi_i \mu_i.\\
&\mbox{s.t.} \quad \psi_j - \varphi_i \le c_{ij},\ \forall i \in \mathcal{I},~\forall j\in \mathcal{J}
\end{aligned}
\end{equation}
Both (\ref{eqn:dis_kp}) and (\ref{eqn:dis_dp}) are linear programming problems, and thus can be solved with generic linear programming methods \cite{karmarkar1984new}, such as dual simplex method.

Then, we introduce the learning the optimal transport in the latent space.

\subsubsection{Training Phase}
Denote by $\{y_j\}_{j \in \mathcal{M}}$ all the $M$ data in the given dataset, where $\mathcal{M}$ is the index set of training samples. Denote by $Enc$ and $Dec$ the pre-trained encoder and decoder on all the given images. First, we use the encoder to get the latent codes of the data $z_{y,j} = Dec(y_j)$. Then, we learn a OT map in the latent space from a simple distribution, unit Gaussian for example, to the empirical distribution formed by $z_{y,j}, j \in \mathcal{J}$. Since the OT map can be computed from the Kantorovich potential, we learn the Kantorovich potential by a two-step manner proposed in \cite{liu2018two}. 
In the first step, we solve Eq. (\ref{eqn:dis_dp}) by solving the following linear programming problem:
\begin{equation}
            \begin{aligned}
            &\max_{H} \quad \frac{1}{m}\sum_{j\in \mathcal{J}} H_j - \frac{1}{m}\sum_{i\in \mathcal{I}} H_i \\
			&\mbox{s.t.}\quad H_j - H_i  \le \frac{1}{2}||z_{x,i} - z_{y,j}||_2^2, \\
            &\qquad \forall i \in \mathcal{I},~\forall j \in \mathcal{J}
\end{aligned}
\label{eqn:lp}
\end{equation}   
where $z_{x,i}$ are sampled from a simple distribution, and $|\mathcal{I}|=|\mathcal{J}|=m$, where $m$ is the batch size. $H$ is the fusion of $H_i$ and $H_j$. $H_i = \varphi_i$, $H_j = \psi_j$, and $\mu_i = \nu_j = 1/m$.

In the second step, we employ a deep neural network $D_w$ parameterized by $w$ to regress the output $H^*$ provided by (14):
\begin{equation}
\label{eq: reg_D}
	\begin{aligned}
	\min_w\mathcal{L}(D_w) = \min_w\frac{1}{m}\sum_{i\in \mathcal{I}} 	(D_w(z_{x,i}) - H^*_i)^2. 
	\end{aligned}
\end{equation}   
However, when the dimensionality of the latent space is high, the number of data is sparse in the high dimension space. Eq. (\ref{eqn:T}) does not necessarily hold given sparse data for computing the Kantorovich potential. Since Eq. (\ref{eqn:T}) is the first order optimum condition of $\min_{x} \varphi(x) + \frac{1}{2} ||x - y||_2^2$, each $x_i = \argmin_{x} \varphi(x) + \frac{1}{2} ||x - y_j||_2^2$ is mapped to $y_j$. Empirically, we can approximate the mapping from $x_i$ to $y_j$ by the following ordering function:
\begin{equation}
\label{eqn: order_org}
\sigma (i) = \argmin_{j\in\mathcal{J}} \frac{1}{2}||x_i-y_j||_2^2 + \varphi_i - \psi_j
\end{equation}
In the latent space, we compute the OT matching from random samples $z_{x,i}, i \in \mathcal{I}$ to the latent representation $z_{y,j}, j \in \mathcal{J}$ of given data using the following ordering function:
\begin{equation}
\label{eqn: order}
\sigma (i) = \argmin_{j\in\mathcal{J}} \frac{1}{2}||z_{x,i}-z_{y,j}||_2^2 + H^*_i - H^*_j
\end{equation}
Instead of optimizing Eq. (\ref{eq: reg_D}), we optimize the following regularized regression problem:
\begin{equation}
\begin{split}
\label{eqn:reg_los1}
\min_{D} \quad \mathcal{L}(D) = \frac{1}{m}\sum_{i\in\mathcal{I}} (D(z_{x,i}) - H^*_i)^2 & \\ +  \frac{\lambda}{m}\sum_{i\in\mathcal{I}} \left (||\nabla_z D(z_{x,i}) || - ||z_{y, \sigma(i)}-z_{x,i} || \right)^2,
\end{split}
\end{equation}
where $\lambda$ is a trade-off parameter. The second term is used to regularize the behavior of $D$'s gradient w.r.t. its input. The total loss ensures that $D$ approximates well in both value and the first order derivative to $\varphi$. 
\subsubsection{Generating Phase}
After we solve (\ref{eqn:reg_los1}), intuitively, $D$ is a smooth approximation of $\varphi$. The OT for a noise $z_x$ is given below using Eq. (\ref{eqn:T}):
\begin{equation}
\label{eqn:T2}
z_y =  z_x + \nabla \varphi(z_x) \approx z_x + \nabla D_z(z_x).
\end{equation}
After we obtain the mapped latent code, we employ the pre-trained decoder to get a generated image $y = Dec(z_y)$.

Figure \ref{fig: workflow} shows the workflow of AE-OT. The Encoder and Decoder are trained and then fixed. In the training phase, the latent representation of a sample $y$ is a real data for the discriminator, while a noise $z$ sampled from a simple distribution is a fake data. We train the discriminator $D$ using the two-step computation  mentioned in this section. In the generating phase, sample a noise $z$ and feed $z + \nabla D(z)$ into the Decoder to produce a generated data $x$. Algorithm \ref{alg:D} and Algorithm \ref{alg:G} present the training and generating phases of AE-OT, respectively.

\begin{algorithm}[t]
  \caption{AE-OT Training Phase}
 \label{alg:D}
  \begin{algorithmic}[1]
\REQUIRE Let $K$ be number of iterations. Batch Size $m$. Adam parameters $\alpha$, $\beta_1$, $\beta_2$. Latent representation $\{z_{y,i} \}_{i \in \mathcal{M}}$ of real data $\{ y_i \}_{i \in \mathcal{M}}$. Regularization parameter $\lambda$.
\ENSURE $D$.
\FOR{$k = 1 \dots K$}
	\STATE Sample $\{z_{y,i} \}_{i \in \mathcal{I}}^{}$ from $\{z_{y,i} \}_{i \in \mathcal{M}}$, where $|\mathcal{I}| = m$.
    \STATE Sample $ \{ z_{x,j}\}_{j \in \mathcal{J}} $ from a simple distribution, where $|\mathcal{J}| = m$.
	\STATE Solve the linear programming problem in (\ref{eqn:lp}).     
%        \STATE Center $\{T_i\}$: $T_i \leftarrow T_i - \frac{1}{m}\sum_{i\in\mathcal{I}}T_i$,
        \STATE Calculate loss $\mathcal{L}$($D$) via Eq. (\ref{eqn:reg_los1}),        
        \STATE Calculate the gradient of $D$: $g_w = \nabla_w\mathcal{L}(D_w)$,
        \STATE Update $w$ with Adam$(w,g_w,\alpha,\beta_1,\beta_2)$.
        
\ENDFOR
\end{algorithmic}
\end{algorithm}

\begin{algorithm}[tbp]
  \caption{AE-OT Generating Phase}
 \label{alg:G}
  \begin{algorithmic}[1]
\REQUIRE Let $D$ be the trained discriminator. $Dec$ is the pre-trained decoder.  $z_{x}$ is sampled from a simple distribution. 
\ENSURE The generated image $y$.
\STATE Generate new codes: $z_{y} = z_{x} + \nabla_{z} D(z_{x})$.
\STATE Decode generated code to an image: $y = Dec(z_{y})$.

\end{algorithmic}
\end{algorithm}

\section{Experiments}
\begin{figure} [t]
\centering     %%% not \center
\subfigure[]{\label{fig:subfig:AE_OMT_iter_5}\includegraphics[width=70mm]{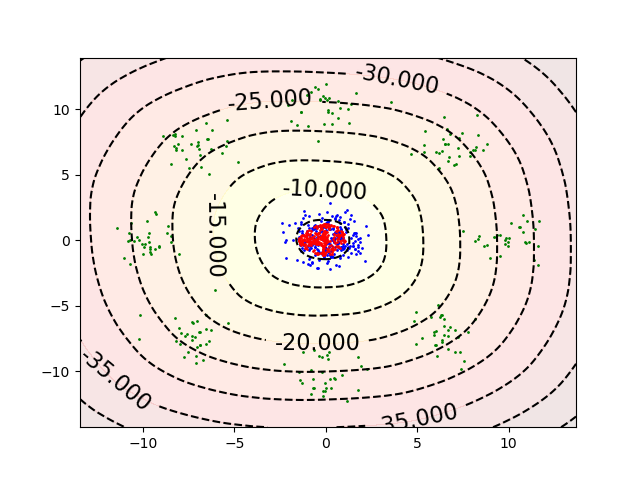}} 
\subfigure[]{\label{fig:subfig:AE_OMT_iter_10000}\includegraphics[width=70mm]{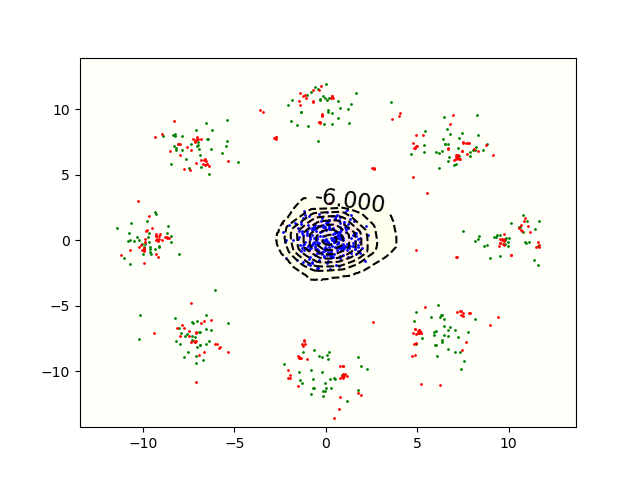}}
\caption{Results on the eight-Gaussian toy dataset. The green points are sampled from the target distribution, and the blue points are sampled form the source distribution. Red points are computed by OT after (a) 5 iterations, and (b) 10000 iterations. Surface values of $D$ are also plotted.}
\label{fig: toy}
\end{figure}
%In the experiments, we evaluate the effectiveness of the proposed method on 
To demonstrate the effectiveness of the proposed method, we 1) evaluate AE-OT on a eight-Gaussian toy dataset; 2) compare AE-OT against VAE \cite{kingma2013auto} and WAE \cite{tolstikhin2017wasserstein} for generative modeling on MNIST \cite{lecun1998gradient} and CelebA \cite{liu2015faceattributes} datasets. AE-OT has a strong geometric motivation and is therefore  more suitable for data that has strong manifold structures, for example, handwritten digits and human faces. 
%Therefore, we conduct experiments on the MNIST and CelebA datasets. 
For WAE, we use the GAN penalty proposed in \cite{tolstikhin2017wasserstein}. In the AE-OT implementation, we set $\lambda=0$ for the eight-Gaussian toy dataset, and $\lambda=0.1$ for the MNIST and CelebA datasets. We use Adam \cite{kingma2014adam} for optimization and set $\beta_1=0.5$ and $\beta_2 = 0.999$ in all the experiments. $\alpha=1e-2$ for the eight-Gaussian experiment; $\alpha = 1e-4$ for the experiments on the MNIST and CelebA datasets.
\begin{figure*} [t]
\centering     %%% not \center
\subfigure[VAE]{\label{fig:subfig:mnist_real}\includegraphics[width=45mm]{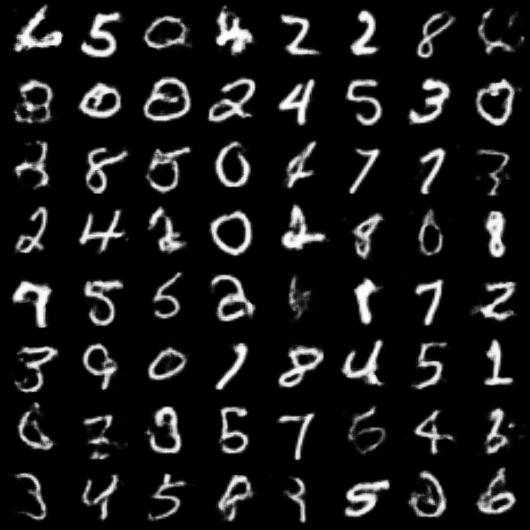}} ~~~~
\subfigure[WAE]{\label{fig:subfig:mnist_WAE}\includegraphics[width=45mm]{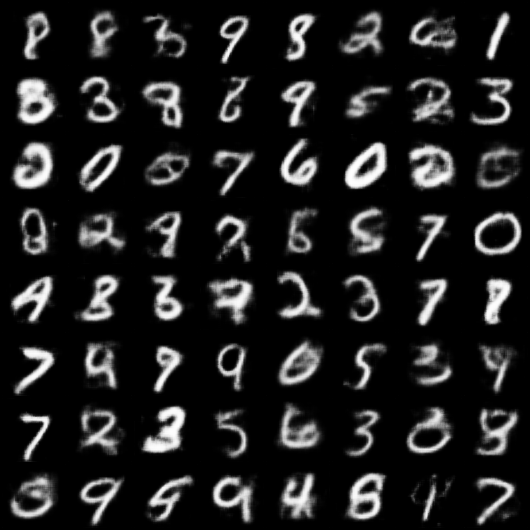}} ~~~~ 
\subfigure[AE-OT]{\label{fig:subfig:mnist_AE_OMT}\includegraphics[width=45mm]{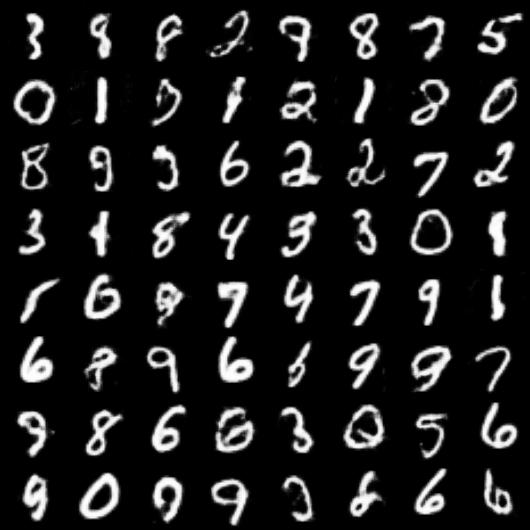}}
\caption{Results on the MNIST dataset. Digits generated by (a) VAE, (b) WAE and (c) AE-OT. All images are randomly generated. }
\label{fig: mnist}
\end{figure*}

\begin{figure*} [htbp]
\centering     %%% not \center
\subfigure[VAE]{\label{fig:subfig:CelebA_VAE}\includegraphics[width=45mm]{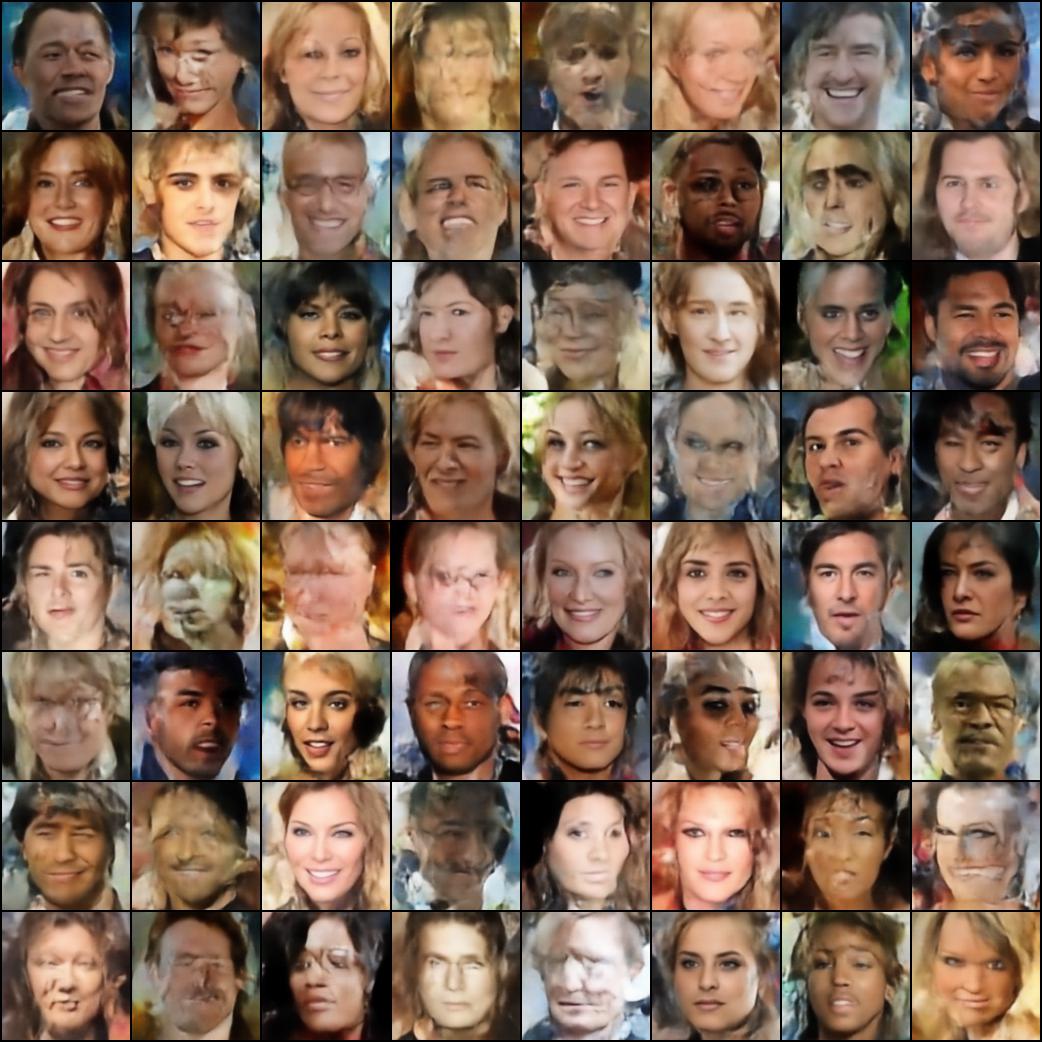}} ~~~~
\subfigure[WAE]{\label{fig:subfig:CelebA_WAE}\includegraphics[width=45mm]{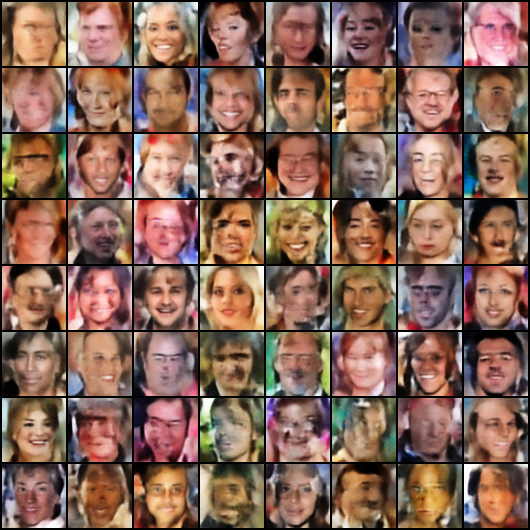}} ~~~~ 
\subfigure[AE-OT]{\label{fig:subfig:CelebA_AE_OMT}\includegraphics[width=45mm]{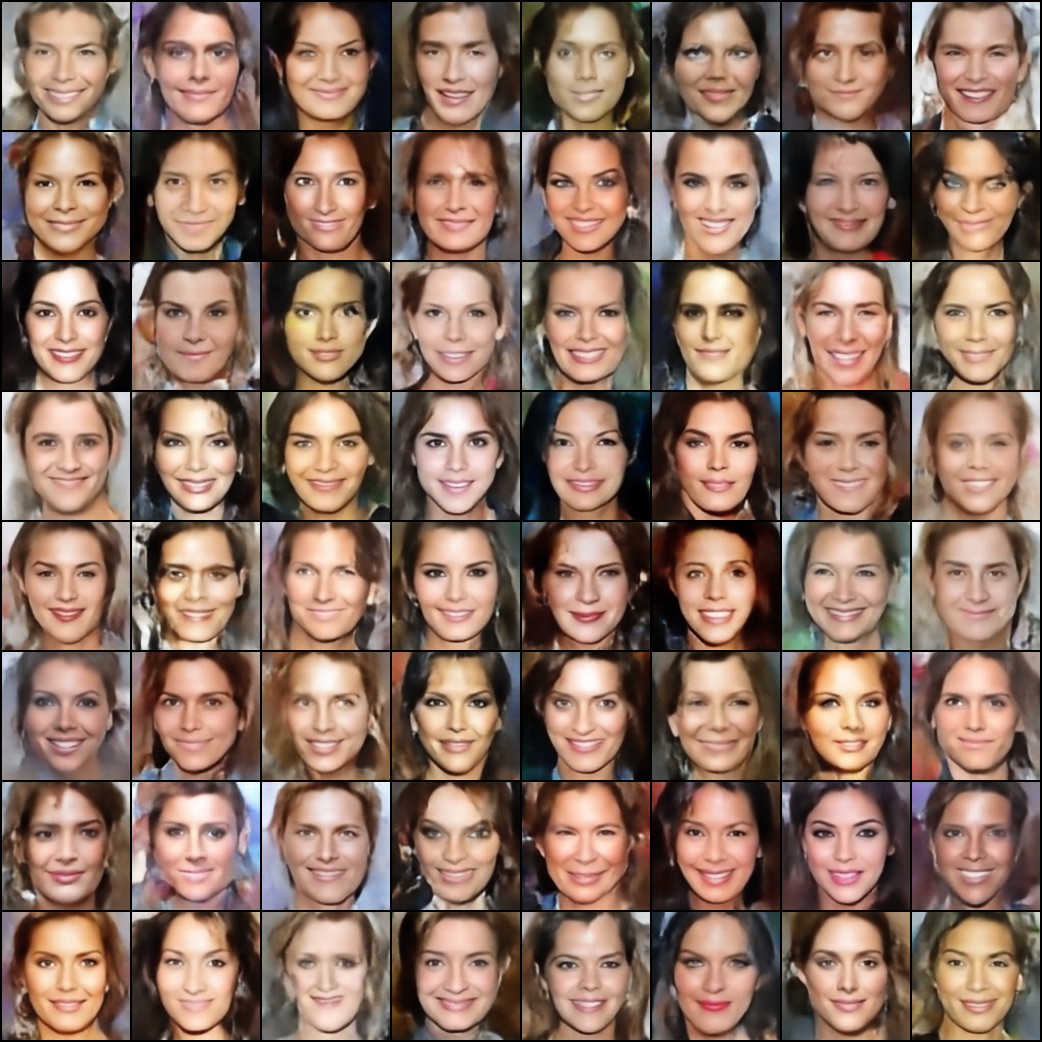}}
\caption{Generated faces on the CelebA dataset by (a) VAE, (b) WAE and (c) AE-OT. All faces are randomly generated.}
\label{fig: celeba}
\end{figure*}

% \begin{figure*} [htbp]
% \centering     %%% not \center
% \subfigure[VAE]{\label{fig:subfig:CelebA_intp_VAE}\includegraphics[width=160mm]{VAE_65_interpolation.png}} ~~~~
% \subfigure[WAE]{\label{fig:subfig:CelebA_intpWAE}\includegraphics[width=160mm]{WAE_196_interpolation.png}} ~~~~ 
% \subfigure[AE-OT]{\label{fig:subfig:CelebA_intp_AE_OMT}\includegraphics[width=160mm]{AE_OT_196_interpolation.png}}
% \caption{Interpolation of the faces generated by (a) VAE, (b) WAE and (c) AE-OT on the CelebA dataset.}
% \label{fig: celeba_intp}
% \end{figure*}

\begin{figure*} [htbp]
\centering     %%% not \center
\includegraphics[width=150mm]{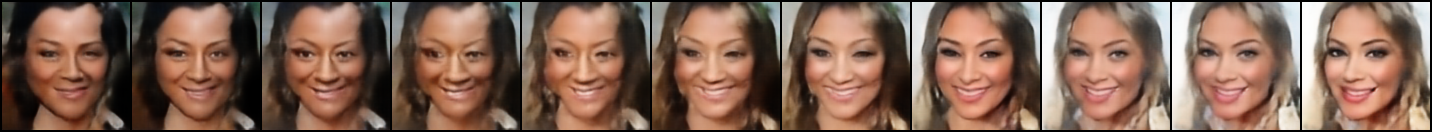}
\caption{Interpolation of the faces generated by AE-OT on the CelebA dataset. }
\label{fig: intp}
\end{figure*}

\textbf{Network Architecture}: For the Auto-Encoder in AE-OT, we use the vanilla Auto-Encoder \cite{jain2017creativity}. For the discriminator of AE-OT on the eight-Gaussian dataset, we use the network architecture used in WGAN-GP \cite{gulrajani2017improved} on the eight-Gaussian dataset. It is a four layer Muti-Layer Perceptions (MLP), with the number of nodes in each hidden layer being 512 and in the output layer being 1. We use ReLU as the non-linear activation function in all layers. For the discriminator of AE-OT on the MNIST and CelebA datasets, we use a six layer MLP, with the number of nodes in each hidden layer being 512 and in the output layer being 1. We use LeakyReLU as the non-linear activation function in all layers and set the slop parameter to 0.2. 

\subsection{Results on the Eight-Guassian Toy Dataset}

\textbf{Dataset Description}: Following previous work \cite{gulrajani2017improved}, we generate a toy dataset consists of eight 2-D Gaussian distributions as the real data distribution. The eight Gaussians are centered at $(10,0)$, $(-10,0)$, $(0,10)$, $(0,-10)$, $(10/\sqrt{2}, 10/\sqrt{2})$, 
$(10/\sqrt{2}, -10/\sqrt{2})$, $(-10/\sqrt{2}, 10/\sqrt{2})$, $(-10/\sqrt{2}, -10/\sqrt{2})$ with $\sigma^2 = 1$. From each Gaussian we sample 32 data points. Therefore the dataset consists of 256 2-D points to represent the real data distribution. The synthetic data is sampled from a Gaussian distribution centered at $(0,0)$ with $\sigma^2 = 1$, from which we sample 256 synthetic data points.

On this 2-D toy dataset, we do not use AE for dimensionality reduction. We train a discriminator using our proposed OT, and use the discriminator to map the synthetic data points to a set of new data points. 
We evaluate whether the transformed new data points form a distribution that is similar to the real data distribution. 
The surface values of the discriminator are plotted in Figure \ref{fig: toy} (a) and (b) after 5 and 10000 discriminator iterations, respectively. The blue points in Figure \ref{fig: toy} (a) and (b) are synthetic data points, while the green points are the real data points. The red points are computed from the synthetic data points using the proposed OT. In Figure \ref{fig: toy} (a), we can see that the generated empirical distribution (red points) is still very close to synthetic distribution since the discriminator has only updated 5 times. After 10000 number of discriminator iterations, the red points form the distribution analogous to the real data distribution. This shows that even though we do not use a generator to generate synthetic samples, with only the discriminator, the source distribution can be transformed to the target distribution. Also, we do not use the regularization term in our method. Computing the Kantorovich potential only on the synthetic data points gives the accurate transport map to transform synthetic data to real data. From this experiment we can see that the proposed method can model the multi-cluster distributions, which is considered as a difficult task for generative models.

\subsection{Results on the MNIST dataset}

\begin{table}
\caption{Inception scores on the MNIST dataset.}
\label{score-tb}
\vskip 0.15in
\begin{center}
\begin{small}
\begin{sc}
\begin{tabular}{lccccr}
\hline

Method & MNIST \\
\hline

VAE    & 1.76 $\pm$ 0.11  \\
WAE    & 1.64 $\pm$ 0.09 \\
AE-OT    & \textbf{1.78} $\pm$ \textbf{0.13} \\
\hline
\end{tabular}
\end{sc}
\end{small}
\end{center}
\vskip -0.1in
\end{table}
In this subsection, we compare our method against VAE and WAE on the MNIST dataset. The images are resized to $64\times64$. For VAE, we use the vanilla Auto-Encoder. The dimensionality of the latent space is set to 10 for all the methods. For VAE, WAE and AE in AE-OT, we train 1000 epochs on the MNIST dataset. For OT in AE-OT, we perform 200K iterations. 

In all the methods, we randomly sample 64 noises and feed them into different generative models. The images generated by different methods are shown in Figure \ref{fig: mnist}. From Figure \ref{fig: mnist} (a) we can see that the brightness of the digits generated by VAE are lower compared to AE-OT. There are some digits that are unclear and incomplete. Figure \ref{fig: mnist} (b) shows digits generated by WAE. Many images are blurry, mainly because WAE trains three networks simultaneously, and thus it cannot learn a desired reconstruction network. Digits produced by AE-OT are visually better than those by VAE and WAE. In addition, we list the Inception Scores (IS) \cite{salimans2016improved} for different methods in Table \ref{score-tb}. From this table we can see that the highest IS is achieved by AE-OT. This experiment shows that the AE-OT is better than VAE and WAE. The main reason is that AE-OT preserves the manifold structure of the data in the latent space.

\subsection{Results on the CelebA dataset}

We compare our method against VAE and WAE on the CelebA dataset. The images are cropped to 128$\times$128. The dimensionality of the latent space is set to 100 for all the methods. We train VAE and AE in AE-OT for 300 epochs on this dataset. For OT in AE-OT, we perform 200K iterations. Our experiment with WAE on this dataset crashes during training. The results shown by WAE are generated just before it crashes.

We sample 64 random noises in the latent space to generate images for all the methods. The generated faces are shown in Figure \ref{fig: celeba}. From this figure we can see that many faces generated by VAE are distorted. VAE tends to mix the face and the background, because VAE enforces the distribution of latent representation of the face manifold to be a unit Gaussian. This distorts the intrinsic representation of the face manifold. Images generated by WAE shown in Figure \ref{fig: celeba} (b) are very blurry and many faces are incomplete. WAE crashes because it jointly trains the encoder, decoder and the discriminator in the latent space. There are competitions among the three networks thus making the training of WAE unstable. In contrast, faces produced by AE-OT (Figure \ref{fig: celeba} (c)) are visually much better than VAE and WAE. The generated face images are clear, complete and recognizable. The faces and the background are well separated, thanks to the property of AE-OT, with which the structure of the latent representation of the face manifold is preserved. 

%In order to verify that the latent space learned by different methods is smooth, we randomly sample two noises in the latent space, interpolate between them and apply different methods to generate faces. Figure \ref{fig: celeba_intp} shows the interpolation of the generated faces. 
% In addition, to examine the manifoldness of the latent space, we conduct latent representation interpolation on the generative model learned by VAE, WAE, and AE-OT(Figure \ref{fig: celeba_intp}). From the results we can see that the interpolation of AE-OT preserves clear facial structure and smooth transition of facial appearance(Figure \ref{fig: celeba_intp}-(c)) while the interpolation obtained by VAE and WAE fall off the face image manifold(Figure \ref{fig: celeba_intp}-(a)(b)).

In order to verify that the latent space learned by AE-OT is smooth, we randomly sample two noises, map them into the latent space using OT. We interpolate between the mapped vectors and forward them into the decoder to generate faces. Figure \ref{fig: intp} shows the interpolation of the faces generated by AE-OT. From this figure we can see that the interpolation of the faces generated by AE-OT preserves clear facial structure and smooth transition of facial appearance. This manifests that the face manifold in the latent space is well preserved by AE-OT. 

%In the middle of Figure \ref{fig: celeba_intp} (a) the faces are nearly disappear. This means that VAE does not model the face manifold well. Faces generated by WAE in Figure \ref{fig: celeba_intp} (b) changes smoothly from left to right, but the face images are very blurry and face qualities are becoming worse and worse. AE-OT produces good face interpolations in Figure \ref{fig: celeba_intp} (c). The faces are very clear compared to those generated by VAE and WAE. The left-most face is gradually changing to the right-most face, and every interpolated face is like a real face. This manifests that the face manifold in the latent space is well preserved by AE-OT. 

\section{Conclusion}
In this work, we propose a novel generative model named AE-OT. Instead of enforcing the distribution of data in the latent space to a simple distribution as proposed in VAE, which leads to the posterior collapse problem. AE-OT transforms a simple distribution to the data distribution in the latent space. In this way, the manifold structure of data in the latent space is preserved, and the posterior collapse problem is addressed. Moreover, In order to avoid the Min-Max optimization problem in a GAN, we propose an OT to generate data from a well-trained discriminator. AE-OT computes the optimal transport map directly in the latent space with explicitly theoretic interpretation. Results on the eight-Gaussian dataset show that the learned OT is capable of handling multi-cluster distributions. Qualitative and quantitative results on the MNIST dataset show that the AE-OT generates better digits than VAE and WAE. Results on the CelebA dataset show that the AE-OT generates much better faces than VAE and WAE, and preserves the manifold structure in the latent space. 

In  future work, we will try to solve the optimal transport more accurately.

\bibliographystyle{abbrv}
\bibliography{ref}

\end{document}